\documentclass[conference]{IEEEtran}
\IEEEoverridecommandlockouts
\usepackage{cite}
\usepackage{amsmath,amssymb,amsfonts}
\usepackage{algorithmic}
\usepackage{graphicx}
\usepackage{textcomp}
\usepackage{xcolor}
\usepackage{subfig}
\def\BibTeX{{\rm B\kern-.05em{\sc i\kern-.025em b}\kern-.08em
    T\kern-.1667em\lower.7ex\hbox{E}\kern-.125emX}}
\usepackage{tikz}
\usepackage{textcomp}
\usepackage[doipre={DOI:~}]{uri}
\usepackage{lipsum}
\newcommand\copyrighttext{%
  \footnotesize \textcopyright 2022 IEEE. Personal use of this material is permitted.  Permission from IEEE must be obtained for all other uses, in any current or future media, including reprinting/republishing this material for advertising or promotional purposes, creating new collective works, for resale or redistribution to servers or lists, or reuse of any copyrighted component of this work in other works. 
  
  Published as a conference paper at the IEEE 2020 ICECS Conference.
  ~\doi{10.1109/ICECS49266.2020.9294863} }
\newcommand{\copyrightnotice}{%
\begin{tikzpicture}[remember picture,overlay]
\node[anchor=south,yshift=10pt] at (current page.south) {\fbox{\parbox{\dimexpr\textwidth-\fboxsep-\fboxrule\relax}{\copyrighttext}}};
\end{tikzpicture}%
}
\begin{document}

\title{Energy-Efficient Adaptive Machine Learning on IoT End-Nodes With Class-Dependent Confidence}

\author{\IEEEauthorblockN{Francesco Daghero\IEEEauthorrefmark{1}, Alessio Burrello\IEEEauthorrefmark{2}, Daniele Jahier Pagliari\IEEEauthorrefmark{1}, Luca Benini\IEEEauthorrefmark{2}\IEEEauthorrefmark{3}, Enrico Macii\IEEEauthorrefmark{4}, Massimo Poncino\IEEEauthorrefmark{1}}
\IEEEauthorblockA{
\IEEEauthorrefmark{1}Department of Control and Computer Engineering, Politecnico di Torino, Italy - name.surname@polito.it}
\IEEEauthorblockA{
\IEEEauthorrefmark{4}Interuniversity Department
of Regional and Urban Studies and Planning, Politecnico di Torino, Italy - enrico.macii@polito.it}
\IEEEauthorblockA{
\IEEEauthorrefmark{2}Energy-Efficient Embedded Systems Laboratory, University of Bologna, Italy - name.surname@unibo.it}
\IEEEauthorblockA{
\IEEEauthorrefmark{3}Integrated Systems Laboratory, ETH Zurich, Switzerland - benini@iis.ee.ethz.ch}
}

\maketitle
\copyrightnotice

\begin{abstract}
Energy-efficient machine learning models that can run directly on edge devices are of great interest in IoT applications, as they can reduce network pressure and response latency, and improve privacy.
An effective way to obtain energy-efficiency with small accuracy drops is to sequentially execute a set of increasingly complex models, early-stopping the procedure for ``easy'' inputs that can be confidently classified by the smallest models. As a stopping criterion, 
current methods employ a single threshold on the output probabilities produced by each model.
In this work, we show that such a criterion is sub-optimal for datasets that include classes of different complexity, and we demonstrate a more general approach based on per-classes thresholds.
With experiments on a low-power end-node, we show that our method can significantly reduce the energy consumption compared to the single-threshold approach.
\end{abstract}

\section{Introduction}
Running Machine Learning (ML) inference directly on IoT edge devices can yield benefits in faster response times, improved data privacy, and higher energy efficiency, by avoiding the transmission of raw sensor data through energy-hungry wireless channels~\cite{chen2019deep,lai2018enabling}.
However, edge inference requires specific optimizations to make complex ML models manageable by battery-operated edge devices with limited computing power. Hardware accelerators achieve impressive efficiency but are only affordable for high-budget and high-volume products~\cite{chen2019deep}. In all other cases, the inference has to be performed on standard microcontrollers (MCUs). Researchers have investigated optimizations for ML inference on MCUs, such as quantization~\cite{choi2018pact,jahierpagliari2016}, and efficient software implementations of the most critical computational kernels~\cite{garofalo2020pulp,lai2018enabling}.

In parallel, platform-independent optimizations of ML models that simplify their execution on constrained devices have also been proposed. In particular, one recent trend is based on tuning the complexity of the inference at runtime, based on the difficulty of the processed input~\cite{Park2015,Tann2016,JahierPagliari2018a, Panda2016,JahierPagliari2019}. The idea is that when inputs are not all equally difficult to process, using a single ML model would either yield an unnecessary complexity for ``easy'' inputs or an accuracy loss for ``difficult'' ones. Therefore, these approaches resort to an \textit{adaptive} inference, where multiple models are used in different combinations depending on the input.

All solutions of this kind use variants of the same policy to determine the classification \textit{confidence} of each model, and consequently, which subset to execute for a given input. Specifically, they impose a \textit{global threshold} on the so-called Score Margin (SM), i.e. the difference between the two largest class probabilities produced in output by a model~\cite{Park2015}. However, this metric is only effective in identifying ``difficult'' inputs for datasets in which all classes have similar complexity.

In this work, we show that a global SM is sub-optimal when, instead, classes are not all equally difficult to process, and that superior results can be obtained using a different threshold per-class. We then propose a methodology for setting such class-dependent thresholds given a desired balance between energy consumption and accuracy.
With experiments on different datasets, we show that our method is able to reduce the energy consumption of 10-60\% compared to a single-threshold approach for the same accuracy level.

\section{Background and Related Work}\label{sec:background}
\begin{figure}[!ht]%
\centering%
\includegraphics[width=.75\columnwidth]{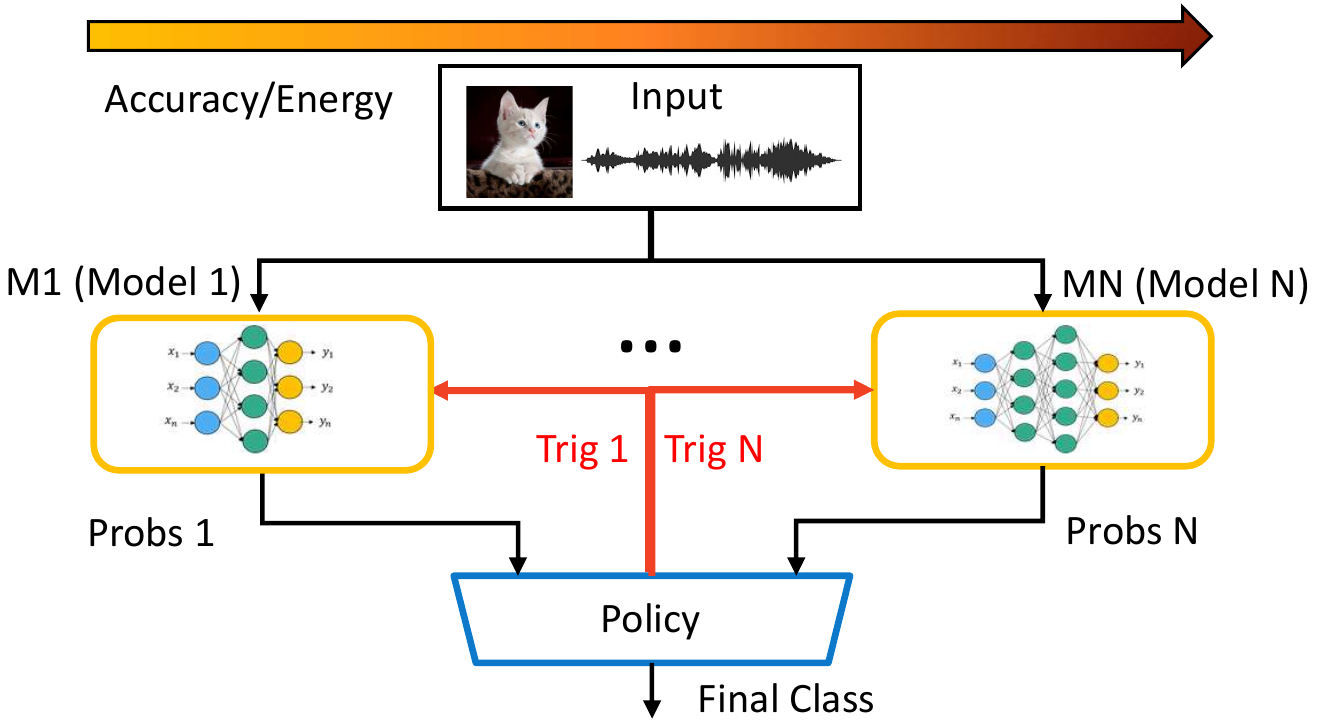}%
\caption{Generic scheme of energy-driven adaptive inference.}\label{fig:method}%
\vspace{-0.2cm}
\end{figure}

We target a family of ``adaptive'' methods for energy-efficient inference, whose generic block diagram is shown in Figure~\ref{fig:method} for a classification task.
A set of ML models ($M1$-$MN$)
are sorted by increasing complexity (and corresponding accuracy)
and sequentially executed in that order.
After the execution of each model, the block labeled \textit{policy} measures its prediction confidence to decide whether to end the classification (for ``easy'' inputs) or to continue with the next model.

Pioneers of this field, the authors of~\cite{Park2015} proposed the so called big/little DNNs, where the models are two convolutional neural networks (CNNs) of increasing size.
Following works have improved this idea avoiding the use of separate DNNs to reduce the overall memory footprint, and extending the technique to $N > 2$.
In~\cite{Tann2016}, ``little'' DNNs are constructed eliminating some channels from the ``big'' one, thus reusing common weights, whereas in~\cite{JahierPagliari2018a} they are obtained progressively decreasing the quantization bit-width. 
In~\cite{Panda2016}, instead, increasingly complex DNNs are constructed using different sets of layers from a single ``big'' model.

These approaches typically measure confidence
as the difference between the first and second largest probabilities produced by the model (so-called Score Margin - SM)~\cite{Park2015,Tann2016,JahierPagliari2018a,Panda2016, JahierPagliari2019}. 
As an example, an adaptive inference method with $N=2$ models generates the following prediction for input $i$:
\begin{equation}\label{eq:ensemble}
y = 
    \begin{cases}
   M1(i)\ if\ \text{SM}(M1(i))>th\\
   M2(i)\ if\ \text{SM}(M1(i)) \leq th
   \end{cases}
\end{equation}
where $th$ is a SM threshold.
The energy consumption of the entire system depends on the complexity of the two models and on the confidence of M1 predictions:
\begin{equation}\label{eq:energy1}
E_{tot} = E(M1) + E(M2) * \text{P}[ \text{SM}(M1(i)) \leq th ] 
\end{equation}
where P indicates probability. Clearly, if the policy always invokes both models, the system reaches the same accuracy of M2 but with an energy overhead, due to running 2 models instead of one. If M2 is never called, instead, the accuracy becomes equal to M1. Therefore, the key element of this method is a reliable confidence estimator, able to distinguish easy from difficult inputs.

In the rest of the paper, we focus on systems with $N=2$ classifiers, such as the big/little DNNs in~\cite{Park2015,JahierPagliari2018a}, to simplify the notation and the corresponding considerations. However, the same approach can be extended to $N > 2$, and this will be the subject of our future work.

\section{Class-Dependent Confidence Estimation}\label{sec:method}
\begin{figure}[h]%
\centering%
\includegraphics[width=0.18\textwidth]{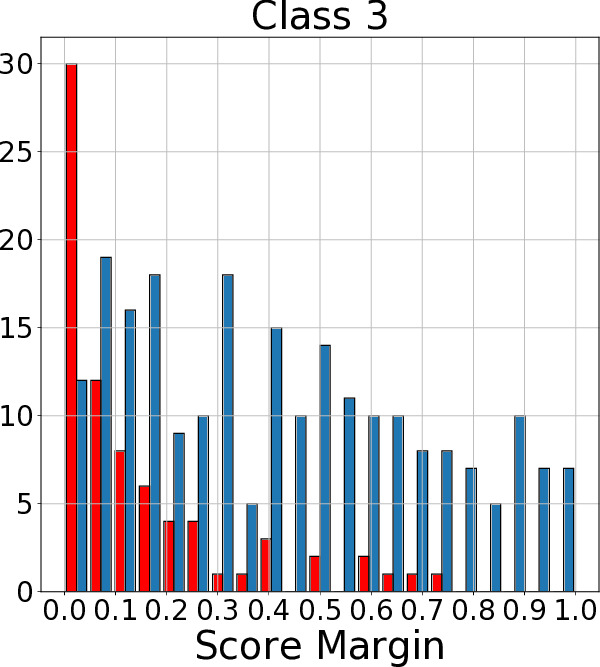}%
\hspace{0.2cm}
\includegraphics[width=0.18\textwidth]{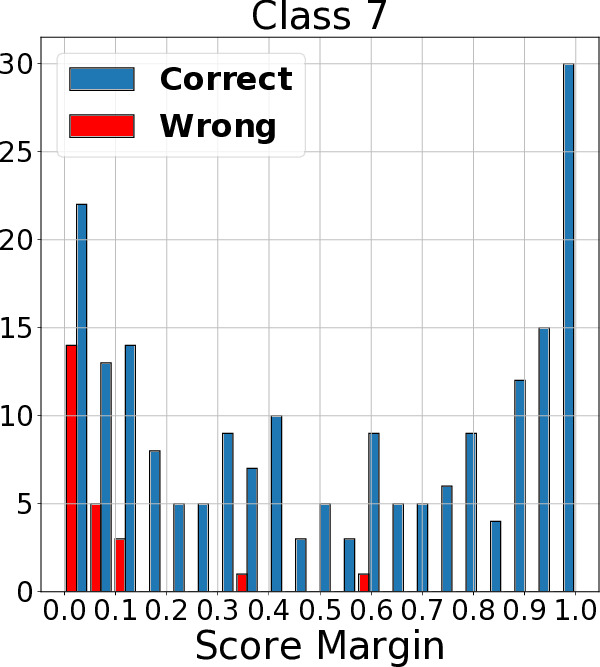}%
\caption{Example of class-dependent SM distribution.}\label{fig:sm_histograms}%
\vspace{-0.3cm}
\end{figure}

All methods described in Section~\ref{sec:background} use a \textit{single} SM threshold in (\ref{eq:ensemble}), regardless of the class. This corresponds to implicitly assuming that `small' models are equally accurate in processing inputs from all classes.
However, for many datasets, this assumption does not hold, as shown in Figure~\ref{fig:sm_histograms}. The histograms show the SM distribution for all validation set samples of the GTSRB dataset~\cite{Stallkamp2012} that, when processed with a logistic regressor (i.e. a single-layer NN), are predicted as belonging to classes 3 and 7. The blue (red) histogram corresponds to samples that are classified correctly (incorrectly) by the model. The classifier and dataset are described in detail in Section~\ref{sec:results}.

The figure clearly shows why a single SM threshold is sub-optimal. When the predicted class is 7, an SM threshold of $th = 0.15$ would be sufficient to correctly invoke the next bigger model for most of the (few) inputs that are wrongly classified, while avoiding further useless processing for most correctly classified samples.
In contrast, the same $th$ applied to class 3 would significantly degrade the accuracy, assuming as correct many wrong classifications. In other words, when this classifier predicts that an input belongs to class 7, the prediction should be assumed correct with high confidence, while inputs predicted as class 3 should be considered with skepticism.
This corresponds to designing a policy able to stop the classification often when the top class is 7 (even if the SM is not so high) and less often when the top class is 3 (only when the SM is close to 1), thus avoiding energy wastes or accuracy losses.
Clearly, the shape of a SM distribution depends both on the nature of the data and on the selected classifier, but these kinds of differences are inevitable when classes have different inherent complexity.

Therefore, we propose a new early-stopping policy that uses class-dependent thresholds $th_c$, optimized as hyper-parameters on the validation set. Specifically, we find the value of $th_c$ for each class $c$ as follows:
\begin{equation}\label{eq:of}
    th_c = \arg\min_{th_c} (FP_c(th_c) + \alpha E_c(th_c)),\ \forall c
\end{equation}
The two addends in (\ref{eq:of}) measure the accuracy of the classification and the energy consumption of the overall system for class $c$ respectively. In particular, $FP_c(th_c)$ is the number of \textit{false positives} generated by the system for class $c$, i.e.:
\begin{equation}\label{eq:fp}
    FP_c(th_c) = \sum_{i: true(i) \neq c}^{M_c} (SM1(i)) > th_c \lor M2(i) \neq true(i))
\end{equation}
where $M_c$ is the number of inputs for which the ``little'' model predicts class $c$, $true(i)$ is the true label of input $i$, and $SM1(i)$ is the score margin for $i$ computed using the probabilities of the ``little'' model $M1$.

As formalized in (\ref{eq:energy1}), in a system with two models, the number of invocations of the ``big'' model is proportional to the energy consumption. Therefore we use:
\begin{equation}
E_c(th_c) = \sum_{i=1}^{M_c} SM1(i) \leq th_c
\end{equation}

The additional factor $\alpha$ in (\ref{eq:of}) is used to balance energy and accuracy in the optimization.
For example, $\alpha=1$ corresponds to giving equal importance to energy and accuracy.
Each single $\alpha$ value yields a corresponding set of $th_c$ (one per class). Exactly as with the standard SM method, users can switch between these sets at runtime, for example giving more importance to energy saving when the battery is low.

\begin{figure}[h]%
\centering%
\includegraphics[width=0.23\textwidth]{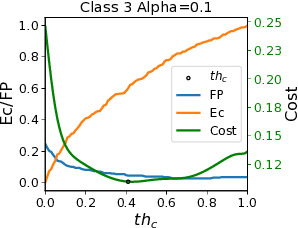}%
\hspace{0.1cm}
\includegraphics[width=0.23\textwidth]{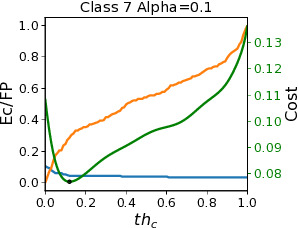}%
\hfill\vspace{0.2cm}%
\includegraphics[width=0.23\textwidth]{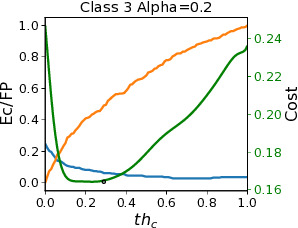}%
\hspace{0.1cm}
\includegraphics[width=0.23\textwidth]{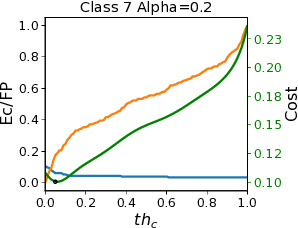}%
\vspace{-0.4cm}
\caption{Objective function for a LeNet-5-like model, for two classes of the GTSRB dataset~\cite{Stallkamp2012} and two values of $\alpha$. }\label{fig:of}%
\vspace{-0.4cm}
\end{figure}

Figure~\ref{fig:of} shows the objective function (\ref{eq:of}) and its two addends as a function of $th_c$ for the same classes of Figure~\ref{fig:sm_histograms} and for two values of $\alpha$. A black dot highlights the minimums. As expected, for a given value of $\alpha$, our method selects a smaller $th_c$ for the easier class 7. Moreover, increasing $\alpha$, i.e. giving more importance to energy reduction, shifts the thresholds for both classes towards smaller values, corresponding to less frequent executions of the ``big'' model.

The minimization of (\ref{eq:of}) is performed offline, and requires a \textit{single} inference on the validation data with each model in the ensemble, storing the corresponding output probabilities and correct labels. Then, the optimal $th_c$ can be obtained with any minimization method (even a grid search).
At runtime, the edge device only needs to store the pre-computed array of $th_c$ corresponding to each desired value of $\alpha$. So, the policy has a negligible impact in terms of both memory occupation and execution time, i.e. a single SM computation and comparison with a threshold, exactly as in the single-threshold approach.

Importantly, if all classes in the dataset have a similar difficulty, hence similar SM distribution, our method simply reduces to the single-threshold approach. In fact, the shape of (\ref{eq:of}) for all classes will be very similar, and so will be also the optimum values of $th_c$. Our method can be outperformed by the single-threshold solution only due to random mismatches between validation and test data distributions (which should not occur in ML best practice) since $th_c$s are set based on the former, e.g. if validation data for a given class are very difficult while test data are easy, or vice versa.
Further, our approach is orthogonal to the way in which individual models are built, so it can be used in combination with any of~\cite{Park2015,Tann2016,JahierPagliari2018a,Panda2016}.

\section{Experimental Results}\label{sec:results}
We tested our proposed method on the STM32H743 MCU by STMicroelectronics, based on an ARM Cortex-M7.
All results refer to floating-point classifiers deployed using X-CUBE-AI.
We considered 3 datasets for image classification and speech recognition tasks.
On CIFAR10~\cite{Krizhevsky09learningmultiple}, we used LeNet-5 and MobileNetV1~\cite{howard2017mobilenets} CNNs as the ``little'' and ``big'' classifier respectively.
Moreover, we also targeted the German Traffic Sign Recognition Benchmark (GTSRB)~\cite{Stallkamp2012}, with 60x60 input images, 
using a logistic regressor as the ``little'' classifier and LeNet-5 as the ``big'' one.
Finally, we considered the Google Speech Commands (GSP)~\cite{warden2018speech} dataset, feeding  
32x32 spectrograms for each word to the classifiers, and reducing the number of classes from 30 to 12 as in~\cite{githubGoogleSpeech},  with a ``bin class' for uncommon words. For this benchmark, we used the same ``little'' and ``big'' models as CIFAR-10.
Models were minimally adapted with respect to the original architectures, changing the first and last layer to match the input size and classes of each dataset.
All results are obtained on test sets, while $th_c$ arrays are optimized on validation sets.

\subsection{Energy versus accuracy trade-off}

\begin{figure}[!ht]%
\vspace{-0.5cm}
\centering%
\includegraphics[width=.9\columnwidth]{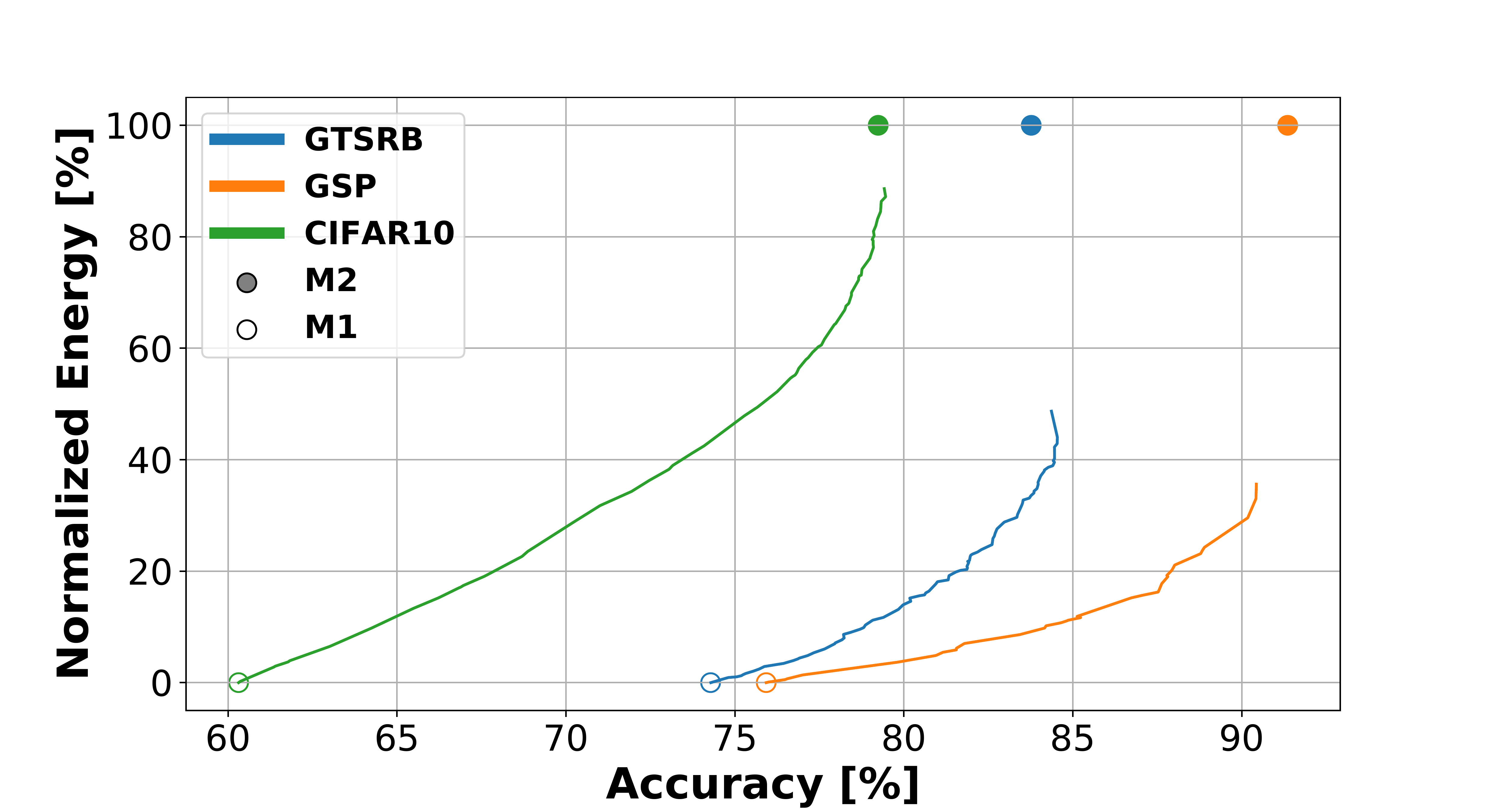}%
\caption{Energy vs accuracy trade-off of the proposed method. Average energy per input normalized to M2 for each dataset.}\label{fig:energyaccuracy}%
\vspace{-0.4cm}
\end{figure}

Figure~\ref{fig:energyaccuracy} shows the trade-off between accuracy and average
energy per input obtained with our method for the three datasets. The three curves are obtained varying $\alpha$,
and the graph also reports the results obtained by M1 and M2 when used individually (dots).
\begin{table}[!ht]%
\centering%
\footnotesize
\caption{Maximum accuracy and energy for the proposed method. Relative differences w.r.t. M2 alone in brackets.}\label{tab:energyaccuracy}%
\begin{tabular}{l|l|l}
                 & \begin{tabular}[c]{@{}l@{}}Accuracy   {[}\%{]}\end{tabular} & \begin{tabular}[c]{@{}l@{}}Energy   {[}mJ{]}\end{tabular} \\ \hline
CIFAR10 {[}15{]} & 79.46 {[}+0.22{]} & 98.76 {[}- 11.4\%{]}  \\
GTSRB {[}17{]}   & 84.54 {[}+0.76{]} & 10.98 {[}- 56.0\%{]} \\
GSP {[}16{]}     & 90.43 {[}-1.01{]} & 40.01 {[}- 64.4\%{]} \\\hline
\end{tabular}
\end{table}
Table~\ref{tab:energyaccuracy} reports the maximum accuracy obtained with the
proposed approach on the three datasets, and the average energy per
input needed to obtain that accuracy, measured on the target hardware platform.
As shown, an accuracy comparable (and sometimes superior) to that of the ``big'' model is reached without invoking M2 for all data, thus significantly reducing the average energy consumption per input. Depending on the dataset, these savings range from 11\% to more than 60\%.

\subsection{Comparison with single-threshold methods}

\begin{table*}
\centering
\footnotesize
\begin{tabular}{l|llll|l}
        & \multicolumn{4}{l|}{Energy [mJ] @ Normalized accuracy gain w.r.t. M1}          & Max. reduction    \\\hline
                  & 25\%             & 50\%           & 75\%           & 100\%           & Acc/Energy  \\\hline
CIFAR10 & 12.37 [-9.6\%]  & 27.46 [-4.5\%] & 46.76 [-2.0\%] & 94.26 [-3.2\%]  &               63.00/8.98 [-14.1\%]\\
GTSRB   & 2.85 [-4.4\%]  & 4.08 [+1.0\%]  & 5.70 [+6.6\%] & 8.98 [-26.4\%] &               84.40/10.05 [-59.3\%]\\
GSP     & 5.44 [-24.1\%] & 10.81 [-14.88\%]  & 17.95 [-12.53\%] & 34.51 [-17.28\%]  &       79.8/5.44 [-24.1\%]  \\ \hline
\end{tabular}
\caption{Energy consumption in different accuracy points. Difference with respect to a single-threshold SM in brackets.}
\label{tab:results_table}
\vspace{-0.3cm}
\end{table*}

Table~\ref{tab:results_table} compares the proposed
method with the single-threshold SM approach of~\cite{Park2015}. It reports the average energy
consumption per input at different fixed trade-off points and the corresponding energy difference (in \%) between our method and the single-threshold one. For example, the column labeled ``25\%'' reports the average energy needed to reach an accuracy that is equal to the accuracy of M1 plus a 25\% of the difference between M2 and M1, while the column labeled ``100\%'' corresponds to the energy needed to reach exactly the accuracy of M2, etc. The column ``Max. reduction'' reports the accuracy condition for which the gain of our method compared to the single-threshold one is maximum. 

Our method reduces the average energy compared to a single-threshold
approach in most accuracy conditions, with gains of 10-60\% depending on the dataset. Results on CIFAR10 are the least impressive, since this dataset contains 10 classes of similar difficulty, while both GSP and GTSRB show more variability.
The few cases where a slight energy increase is obtained can be charged to the difference between validation and test data distributions, as explained in Section~\ref{sec:method}.

\begin{figure}[ht]%
\centering%
\includegraphics[width=.6\columnwidth]{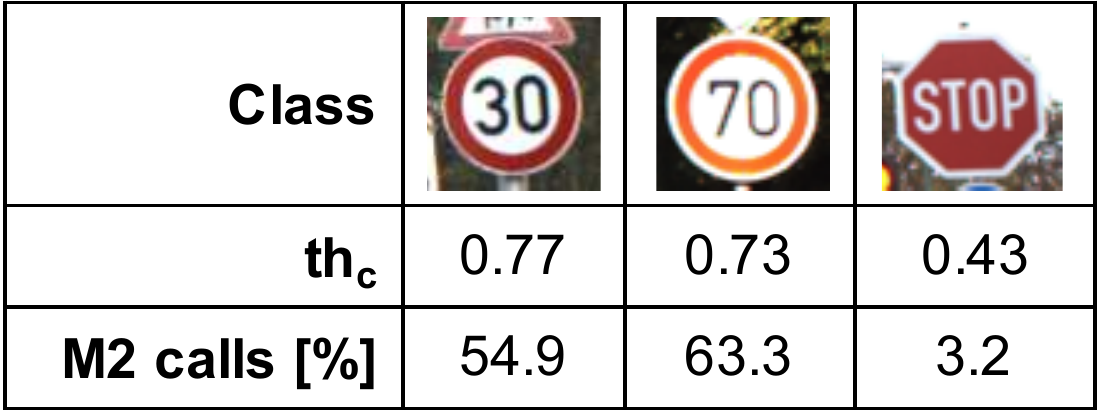}%
\caption{Example of ``difficult'' and ``easy'' classes for the GTSRB dataset. $th_c$ and \% of M2 calls are for $\alpha = 0.05$.}\label{fig:signs_example}%
\end{figure}

Figure~\ref{fig:signs_example} shows a qualitative example (from GTSRB) of the fact that our method tends to assign larger SM thresholds (corresponding to more invocations of the ``big'' model) to difficult classes.
Indeed the two speed limit signs, which could be easily mistaken for one another are assigned higher $th_c$. In contrast, the ``stop'' sign, which is easily recognizable, is assigned a lower $th_c$ to avoid useless ``big'' model executions. Similar considerations can be done for the other datasets.

\subsection{Effect on imbalanced datasets}

Our method is also effective when training and val/test data are differently balanced. This happens, for example, when using a pre-trained model whose training class frequencies are not those expected in the final application (e.g. a generic speech recognition model used to perform wake-word recognition). 
This negatively impacts single-threshold SM methods such as~\cite{Park2015,Tann2016,JahierPagliari2018a}, since classes with low \textit{prior probabilities} yield lower scores, and therefore generate lower SM. A single-threshold approach would wrongly assume that the classifier is not confident about these predictions, and call ``big'' model(s) more often, even if those lower margins are only due to a class being less present in the training set, and not to its difficulty.

\begin{figure}[ht]%
\vspace{-0.4cm}
\centering%
\includegraphics[width=.7\columnwidth]{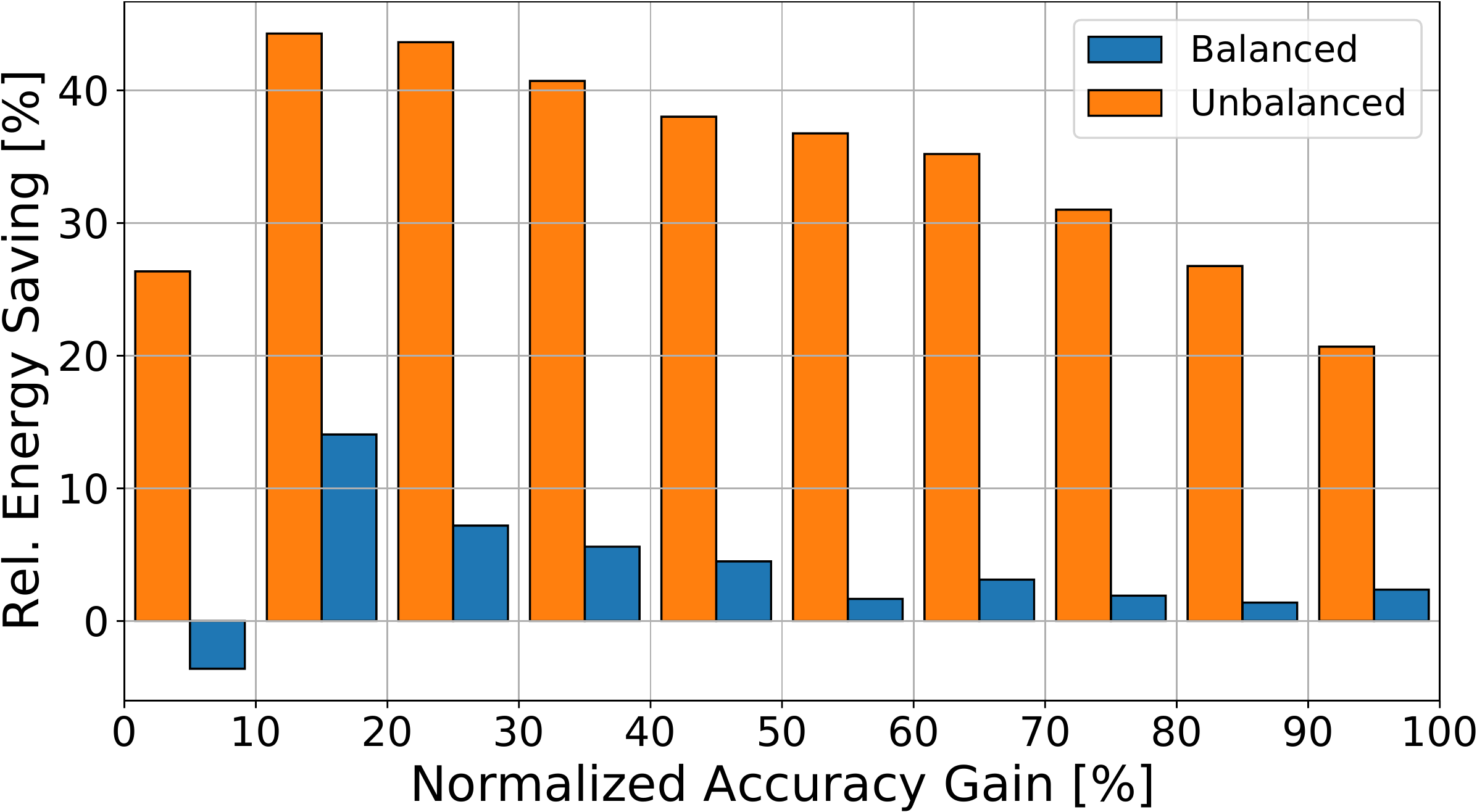}%
\caption{Energy gains of class-dependent SM thresholds on CIFAR10 and its artificially unbalanced version.}\label{fig:unbbal_cifar10}%
\vspace{-0.6cm}
\end{figure}

Class-dependent thresholds can automatically compensate for these priors mismatches, allowing accurate classification without the need of re-training. The only requirement is the availability of a small correctly balanced validation set to optimize $th_c$.
To show this, we have artificially unbalanced the training set of CIFAR10, undersampling 8 random classes to 1/5 of the original images. We have then computed $th_c$ on the (balanced) validation set and evaluated the average energy and accuracy of our method on the test set. Figure~\ref{fig:unbbal_cifar10} shows the percentage energy saving with respect to a single-threshold approach for different accuracy points. While our method yields consistent savings also on the normally balanced CIFAR10, the benefits increase significantly when training and test data are not similarly balanced, reaching more than 40\%.

\section{Conclusions}
We have presented a novel policy to guide the execution of energy-driven adaptive ML inference. Our approach uses class-dependent SM thresholds to estimate the confidence of a prediction, based on the assumption that classes are not all equally difficult to distinguish in most datasets. With experiments on a real edge MCU, we have shown that this approach yields consistent energy savings at iso-accuracy compared to a solution that uses a single SM threshold.


\begin{thebibliography}{10}
\providecommand{\url}[1]{#1}
\csname url@samestyle\endcsname
\providecommand{\newblock}{\relax}
\providecommand{\bibinfo}[2]{#2}
\providecommand{\BIBentrySTDinterwordspacing}{\spaceskip=0pt\relax}
\providecommand{\BIBentryALTinterwordstretchfactor}{4}
\providecommand{\BIBentryALTinterwordspacing}{\spaceskip=\fontdimen2\font plus
\BIBentryALTinterwordstretchfactor\fontdimen3\font minus
  \fontdimen4\font\relax}
\providecommand{\BIBforeignlanguage}[2]{{%
\expandafter\ifx\csname l@#1\endcsname\relax
\typeout{** WARNING: IEEEtran.bst: No hyphenation pattern has been}%
\typeout{** loaded for the language `#1'. Using the pattern for}%
\typeout{** the default language instead.}%
\else
\language=\csname l@#1\endcsname
\fi
#2}}
\providecommand{\BIBdecl}{\relax}
\BIBdecl
\bibitem{chen2019deep} J.~Chen and X.~Ran, ``Deep learning with edge computing: A review,'' \emph{Proc. of the IEEE}, vol. 107, no.~8, pp. 1655--1674, 2019.

\bibitem{lai2018enabling} L.~Lai and N.~Suda, ``Enabling deep learning at the lot edge,'' in \emph{ICCAD}, 2018, pp. 1--6.

\bibitem{choi2018pact} J.~Choi et al, ``Pact: Parameterized clipping activation for quantized neural networks,'' \emph{arXiv preprint arXiv:1805.06085}, 2018.

\bibitem{jahierpagliari2016} D.~Jahier Pagliari et al, ``Energy-efficient
    Digital Processing via Approximate Computing'', Smart Systems Integration and Simulation, pp.55–89, Springer, 2016.

\bibitem{garofalo2020pulp} A.~Garofalo et al, ``Pulp-nn: accelerating quantized neural networks on parallel ultra-low-power risc-v processors,'' \emph{Philos. Trans. R. Society A}, vol. 378, no. 2164, p. 20190155, 2020.

\bibitem{howard2017mobilenets} A.~G. Howard et al, ``Mobilenets: Efficient convolutional neural networks for mobile vision applications,'' \emph{arXiv preprint arXiv:1704.04861}, 2017.

\bibitem{Park2015} E.~Park et al, ``{Big/little deep neural network for ultra low power inference},'' in \emph{CODES+ISSS}, 2015, pp. 124--132.

\bibitem{Tann2016} H.~Tann et al, ``{Runtime configurable deep neural networks for energy-accuracy trade-off},'' in \emph{CODES+ISSS}, 2016, pp. 1--10.

\bibitem{JahierPagliari2018a} D.~{Jahier Pagliari} et al, ``{Dynamic Bit-width Reconfiguration for Energy-Efficient Deep Learning Hardware},'' in \emph{ISLPED}, 2018, pp. 47:1----47:6.

\bibitem{Panda2016} P.~Panda et al, ``{Conditional Deep Learning for Energy-Efficient and Enhanced Pattern Recognition},'' in \emph{DATE}, 2016, pp. 475--480.

\bibitem{JahierPagliari2019} D.~{Jahier Pagliari} et al, ``{Dynamic Beam Width Tuning for Energy-Efficient Recurrent Neural Networks},'' in \emph{GLSVLSI}, 2019, pp. 69--74.

\bibitem{Stallkamp2012} J.~Stallkamp et al, ``Man vs. computer: Benchmarking machine learning algorithms for traffic sign recognition,'' \emph{Neural Networks}, vol.~32, pp. 323--332, 2012.

\bibitem{Krizhevsky09learningmultiple} A.~Krizhevsky, ``Learning multiple layers of features from tiny images,'' Tech. Rep., 2009.

\bibitem{warden2018speech} P.~Warden, ``Speech commands: A dataset for limited-vocabulary speech recognition,'' \emph{arXiv preprint arXiv:1804.03209}, 2018.

\bibitem{githubGoogleSpeech} \url{https://github.com/tugstugi/pytorch-speech-commands}

\end{thebibliography}
\end{document}